%% file: main.tex
\documentclass[10pt,twocolumn,letterpaper]{article}

\usepackage[pagenumbers]{cvpr} % To force page numbers, e.g. for an arXiv version

\usepackage{graphicx}
\usepackage{amsmath}
\usepackage{amssymb}
\usepackage{booktabs}
\usepackage{multirow}
\usepackage{makecell}

\newcommand{\keywords}[1]{}

\newcommand{\red}[1]{#1}
\usepackage[pagebackref,breaklinks,colorlinks]{hyperref}

\usepackage[capitalize]{cleveref}
\crefname{section}{Sec.}{Secs.}
\Crefname{section}{Section}{Sections}
\Crefname{table}{Table}{Tables}
\crefname{table}{Tab.}{Tabs.}

\def\cvprPaperID{02307} % *** Enter the CVPR Paper ID here
\def\confName{ECCV}
\def\confYear{2024}

\begin{document}

%%%%%%%%% TITLE - PLEASE UPDATE
\title{VISA: Reasoning Video Object Segmentation via Large Language Models}

\author{
Cilin Yan\footnotemark[1]\\
Beihang University\\
\and
Haochen Wang\footnotemark[1]\\
University of Amsterdam \\
\and
Shilin Yan,Xiaolong Jiang,Yao Hu\\
Xiaohongshu Inc.\\
\and
Guoliang Kang\footnotemark[2]\\
Beihang University\\
% Zhongguancun Lab.\\
\and
Weidi Xie\\
Shanghai Jiao Tong University\\
% Shanghai AI Lab.\\
\and
Efstratios Gavves\\
University of Amsterdam
}
\maketitle
\renewcommand{\thefootnote}{\fnsymbol{footnote}} %将脚注符号设置为fnsymbol类型，即特殊符号表示
\footnotetext[1]{Equal contribution.}  
\footnotetext[2]{Corresponding author.} %对应脚注[2]

\input{section/0-abstract}
\input{section/1-introduction}
\input{section/2-related_work}
\input{section/3-method}
\input{section/4-experiment}
\input{section/5-conclusion}

% \noindent
% \textbf{Acknowledgement:}
\section*{Acknowledgement}
This project is supported by National Natural Science Foundation of China under Grant 92370114 and European Union (ERC, EVA, 950086).

%%%%%%%%% REFERENCES
{\small
\bibliographystyle{ieee_fullname}
\bibliography{main}
}

\input{section/6-appendix}

\end{document}

%% file: section/0-abstract.tex
\begin{abstract}
Existing Video Object Segmentation~(VOS) relies on explicit user instructions, such as categories, masks, or short phrases, restricting their ability to perform complex video segmentation requiring reasoning with world knowledge.
In this paper, we introduce a new task, Reasoning Video Object Segmentation~(ReasonVOS). 
This task aims to generate a sequence of segmentation masks in response to implicit text queries that require complex reasoning abilities based on world knowledge and video contexts,
which is crucial for structured environment understanding and object-centric interactions, pivotal in the development of embodied AI.
To tackle ReasonVOS, we introduce VISA~(Video-based large language Instructed Segmentation Assistant), to leverage the world knowledge reasoning capabilities of multi-modal LLMs while possessing the ability to segment and track objects in videos with a mask decoder.
Moreover, we establish a comprehensive benchmark consisting of \red{35,074} instruction-mask sequence pairs from \red{1,042} diverse videos, which incorporates complex world knowledge reasoning into segmentation tasks for instruction-tuning and evaluation purposes of ReasonVOS models.
Experiments conducted on 8 datasets demonstrate the effectiveness of VISA in tackling complex reasoning segmentation and vanilla referring segmentation in both video and image domains.
The code and dataset are available at \url{https://github.com/cilinyan/VISA}.

\vspace{-2mm}

\end{abstract}

%% file: section/1-introduction.tex
\section{Introduction}
\label{sec:intro}

Existing video object segmentation relies on explicit queries, such as pre-defined categories~\cite{wang2021end, wu2021seqformer, cheng2021mask2former}, masks of certain frames~\cite{cheng2021rethinking, yang2021associating}, or explicit short phrases describing intuitive features~\cite{seo2020urvos, wu2022language, botach2022end}.
Such systems lack the capacity to reason and infer users’ intentions based on implicit instructions.
For instance, it is more intuitive for the users to give instructions like ``Find my favorite cup'' instead of ``Find the red cup located second to the left on the table''.
To accomplish the first instruction, the model needs to understand that ``favorite'' means ``most frequently used'' to some degree, and callback the history temporal information to localize the cup.

In this work, we propose Reasoning Video Object Segmentation~(ReasonVOS), which aims to generate a binary mask sequence given complex and implicit text instruction in videos.
Notably, the text instruction is not limited to a straightforward reference~(e.g., the running car), but a more complex description including reasoning of world knowledge~(e.g., the car powered by electricity).
This task requires the integration of reasoning ability with long-term video understanding to accurately localize target objects in videos, which is a crucial ability for Embodied AI systems that enable robots to fulfill effective interaction with objects in dynamic environments given user instructions.

\begin{figure*}[!t]
\begin{center}
\includegraphics[width=0.98\linewidth]{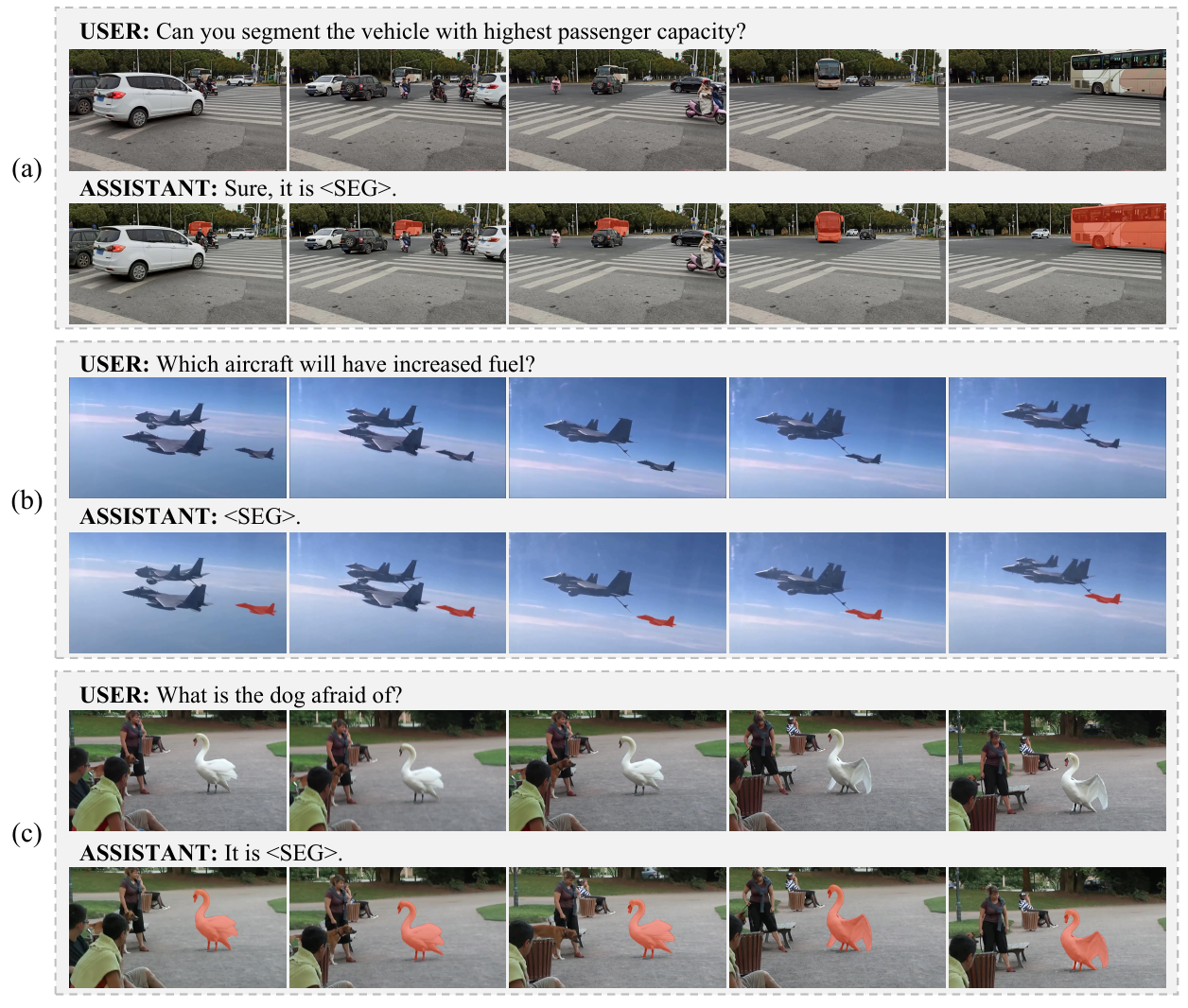}
\end{center}
\vspace{-5mm}
\caption{We enable the reasoning video object segmentation capabilities for current multi-modal LLMs.
The proposed VISA is capable of segmenting and tracking objects given text descriptions involving: (a) complex reasoning of world knowledge; (b) inference of upcoming events; and (c) comprehensive understanding of video content.
}
\vspace{-3mm}
\label{fig:demonstrate}
\end{figure*}

To tackle Reasoning Segmentation in images, recent work LISA~\cite{lai2023lisa} leverages the language generation prowess of multi-modal LLMs, complemented by a mask decoder for generating segmentation results. 
However, the Reasoning Segmentation in videos demands temporal information for comprehensive video understanding and spatial details for producing high-quality segmentation mask sequences.
Therefore, the multi-modal LLMs need to simultaneously process multiple frames with a substantial number of tokens for each frame. 
Considering the numerous visual tokens to be processed simultaneously, it is computationally intractable to directly broadcast an image reasoning segmentation model to the video domain.

To this end, we introduce VISA (Video-based large language Instructed Segmentation Assistant), designed to efficiently encode long-term video features while preserving spatial details to enable reasoning in video object segmentation.
Specifically, we start by designing a Text-guided Frame Sampler~(TFS) to select frames that are most relevant to the task based on textual instructions, focusing the model on the most significant moments for object identification. 
TFS reduces the requirement of visual token numbers to be handled, enabling VISA to process long-term videos. 
These selected frames, along with the text queries, 
are tokenized and processed concurrently by a multi-modal Large Language Model (LLM), enabling sophisticated reasoning over video content and facilitating the generation of precise textual outputs. 
To equip VISA with robust segmentation capabilities, we incorporate a special token $<$\textsc{Seg}$>$ in the output text, inspired by the approach in LISA~\cite{lai2023lisa}. 
The hidden embedding of $<$\textsc{Seg}$>$ is leveraged to produce segmentation masks of selected frames using a SAM~\cite{kirillov2023segment} decoder. 
The segmentation process is completed by deriving the masks for the remaining frames with an object tracker~\cite{cheng2022xmem}.
As illustrated in Fig.~\ref{fig:demonstrate}, VISA demonstrates remarkable proficiency in handling complex segmentation tasks that require: (a) reasoning based on world knowledge; (b) inference of future events; and (c) a comprehensive understanding of video content.

%\weidi{edited version, better to be more structured, something like our model is consisted of $n$? components,
%(i) a text-guided frame selector, that exploits...
%(ii) a .... that ....
%(iii) ..... that....}

\begin{figure*}[!t]
\begin{center}
\includegraphics[width=0.7\linewidth]{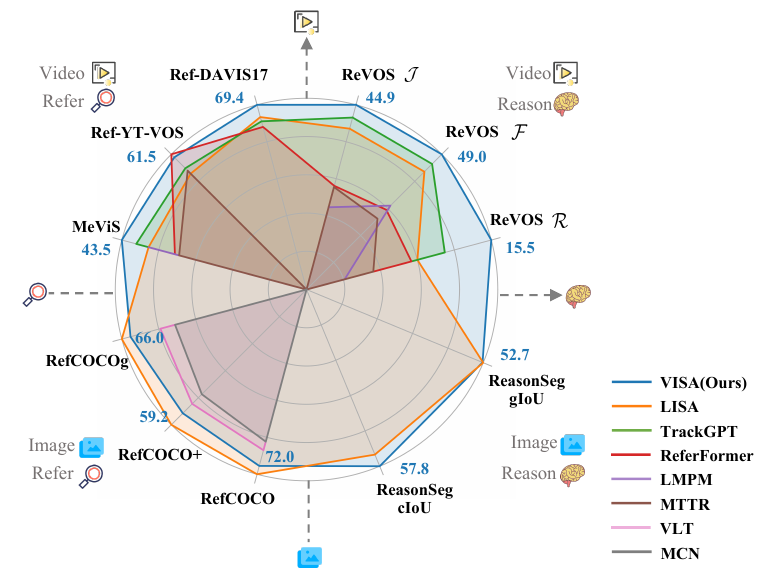}
\end{center}
\vspace{-5mm}
\caption{Our proposed VISA consistently achieves state-of-the-art performances on video and image datasets over reasoning and referring segmentation tasks. $\mathcal{J}$ is region similarity~\cite{seo2020urvos}, $\mathcal{F}$ is contour accuracy~\cite{seo2020urvos}, and $\mathcal{R}$ is robustness score~\cite{li2022r}.
}
\vspace{-3mm}
\label{fig:radar}
\end{figure*}

To evaluate the effectiveness of the proposed VISA, we create a benchmark dataset named ReVOS. 
This dataset comprises \red{35,074} pairs of instruction-mask sequences derived from \red{1,042} diverse videos. In contrast to traditional referring video segmentation datasets, 
such as Ref-YouTube-VOS~\cite{seo2020urvos} and MeViS~\cite{ding2023mevis}, 
which primarily contain explicit short phrases, ReVOS includes text instructions that necessitate a sophisticated understanding of both video content and general world knowledge. We carry out comprehensive experiments on the ReVOS dataset as well as on seven existing segmentation datasets. 
The results in Fig.~\ref{fig:radar} demonstrate that VISA not only facilitates advanced reasoning segmentation in both video and image domains but also achieves competitive performance on referring segmentation tasks.

Our main contributions could be summarized as follows:
(i) We introduce a new task ReasonVOS~(Reasoning Video Object Segmentation), which aims to segment and track objects in videos given implicit texts. 
ReasonVOS emphasizes the requirements of reasoning, summary, and inference ability based on video content and world knowledge, crucial for an intelligent perception system to interact with dynamic environments.
(ii) We propose VISA~(Video-based large language Instructed Segmentation Assistant), which efficiently integrates long-term video features and complex text queries to enable the reasoning video object segmentation ability. 
% (iii) We collect a large-scale dataset ReVOS, comprising 1,042 videos and 35,074 implicit object descriptions for instruction tuning and evaluation purposes of ReasonVOS models. The experiments on ReVOS and existing datasets show that our proposed VISA performs robustly in reasoning segmentation tasks of both image and video domains.
(iii) We collect a large-scale dataset ReVOS, comprising \red{1,042} videos and \red{35,074} object descriptions for instruction tuning and evaluation purposes of ReasonVOS models. The experiments on ReVOS and existing datasets show that our proposed VISA performs robustly in reasoning segmentation tasks of both image and video domains.

%% file: section/2-related_work.tex
\section{Related Work}
\noindent \textbf{Video Object Segmentation.}
Video Object Segmentation~(VOS) is designed to segment and track objects in videos based on specific references, including categories~\cite{cheng2021mask2former, wu2021seqformer, wang2021end,zhang2023dvis,zhang2023dvis++,zhou2024dvis,zhang20231st}, segmentation masks~\cite{cheng2022xmem, perazzi2017learning, cheng2018fast, oh2019video}, or explicit text descriptions~\cite{seo2020urvos,ding2023mevis,botach2022end,8100260,wu2022language}. 
VOS plays a critical role in structured video representation learning and Embodied AI.
Category-based VOS methods~(or Video Instance Segmentation), such as Mask2Former~\cite{cheng2021mask2former}, SeqFormer~\cite{wu2021seqformer}, and VisTR~\cite{wang2021end}, segment and associate objects in videos given a pre-defined category list.
Mask-based VOS methods~(or semi-supervised VOS), such as STM~\cite{oh2019video} and XMem~\cite{cheng2022xmem}, segment and track objects in videos based on the segmentation mask given in certain frames.
The utility of the aforementioned approaches is constrained by their reliance on structured and straightforward input, resulting in limited generalizability in real-world scenarios that necessitate complex reasoning and flexible input formulation.

In contrast, the text-based VOS~(Referring VOS)~\cite{seo2020urvos,ding2023mevis,botach2022end}, aims to segment objects in videos given text description.
However, the text descriptions in Referring VOS fall into short phrases indicating the explicit object information, such as action, localization, and appearance.
This system lacks the ability to handle complex sentences that involve common sense reasoning or inference based on video content.
In this work, we introduce ReasonVOS, extending the short phrases to complex sentences requiring reasoning and the inference of world knowledge alongside video content. This advancement significantly enhances the practical utility of VOS across various tasks.

\noindent \textbf{Multi-Modal Large Language Model.}
Inspired by the impressive reasoning capabilities of Large Language Models (LLMs), researchers are investigating methods to transpose these abilities into the vision domain, leading to the development of multi-modal LLMs~\cite{alayrac2022flamingo,zhang2024omg,wang2023cogvlm}.
Flamingo~\cite{alayrac2022flamingo} utilizes a cross-attention structure to attend to visual contexts, facilitating visual in-context learning. 
Meanwhile, models like BLIP-2~\cite{li2023blip} and mPLUG-OWL~\cite{ye2023mplug} propose the encoding of image features using a visual encoder, which are subsequently integrated into the LLM along with text embeddings.
Otter~\cite{li2023otter} further incorporates robust few-shot capabilities through in-context instruction tuning on the proposed MIMIC-IT dataset. 
LLaVA~\cite{liu2024visual} and MiniGPT-4~\cite{zhu2023minigpt} first conduct image-text feature alignment followed by instruction tuning and also investigate image retrieval for LLMs.

Recent studies have delved into the confluence of multi-modal Large Language Models (LLMs) and vision tasks. VisionLLM~\cite{wang2024visionllm} provides a versatile interface for engaging with various vision-centric tasks through instruction tuning but fails to fully exploit LLMs for complex reasoning. 
Kosmos-2~\cite{peng2023kosmos} builds a large-scale dataset of grounded image-text pairs, thereby injecting grounding capabilities into LLMs. 
DetGPT~\cite{pi2023detgpt} connects the multi-modal LLMs and open-vocabulary detectors, facilitating detection tasks based on user instructions. 
GPT4RoI~\cite{zhang2023gpt4roi} innovates by incorporating spatial boxes as input and training the model on region-text pairings.
LISA~\cite{lai2023lisa} efficiently enables segmentation capabilities of multi-modal LLMs in the image domain by introducing a special $<$\textsc{Seg}$>$ token.
All the above-mentioned methods focus on downstream tasks in the image domain.
The concurrent work TrackGPT~\cite{stroh2024trackgpt}, made the first attempt to tackle reasoning segmentation in videos.
However, TrackGPT processes single frames at one time and segments the objects frame-by-frame without any temporal correspondence, which falls in complex scenarios requiring long-term video understanding.
On the contrary, our proposed VISA handles multiple frames at one time to obtain long-term awareness.

\noindent \textbf{Video Multi-Modal Large Language Model.}
To support video understanding in LLMs, Video-LLaMA~\cite{zhang2023video} attempts to utilize BLIP-2 for video embedding extraction, while Video-ChatGPT~\cite{maaz2023video} proposes spatial and temporal pooling for video features. 
However, given the substantial number of tokens required for each frame, LLMs encounter significant challenges when processing extensive video sequences. 
It prevents previous work~\cite{maaz2023video,zhang2023video} from representing long video sequences that exceed a duration of one hour in LLMs. 
To solve the issue, LLaMA-VID~\cite{li2023llama} proposes to efficiently encode each frame with only 2 tokens, which supports long video understanding in existing LLMs.
Those works either use pooling or projection to abstract each frame into a few visual tokens, which is inadequate to provide detailed spatial information for segmentation.
In this work, we first select significant frames for identifying the target objects and simultaneously process the selected frames with a large number of visual tokens, avoiding the spatial pooling or projection and thus benefiting the segmentation tasks.

%% file: section/3-method.tex
\section{Method}

%figure
\begin{figure*}[t]
\begin{center}
\includegraphics[width=0.98\linewidth]{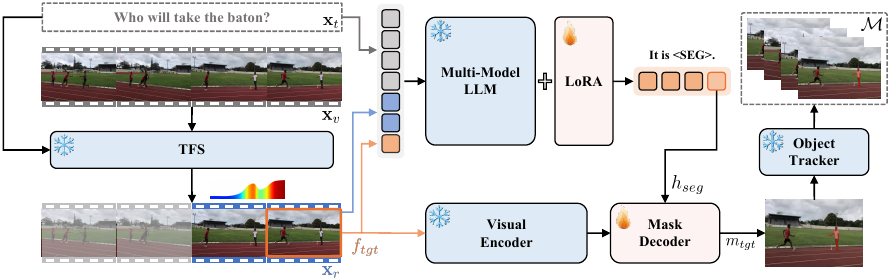}
\end{center}
\vspace{-5mm}
\caption{Overview of VISA. (a) Given a video $\mathbf{x}_v$ and a text description $\mathbf{x}_t$, a Text-guided Frame Sampler~(TFS) is proposed to sample the most distinguishing frame $f_{tgt}$ as the target to be segmented and corresponding reference frames $\mathbf{x}_r$. 
(b) Then $f_{tgt}$, $\mathbf{x}_r$, and $\mathbf{x}_t$ are tokenized and fed to a Multi-Modal LLM to generate text output, including a special token $<$\textsc{Seg}$>$. The last-layer embedding of $<$\textsc{Seg}$>$ token $h_{seg}$ is then decoded into the segmentation mask $m_{tgt}$ of frame $f_{tgt}$ via the mask decoder. (c) Finally, the segmentation masks of all frames $\mathcal{M}$ are generated by propagation with an Object Tracker.
The modules in blue are frozen during the training, while the modules in pink are trainable.}
\label{fig:architecture}
\vspace{-3mm}
\end{figure*}

\subsection{Task Setting}

In this section, we start by defining the task of interest, termed ReasonVOS.
Specifically, given a high-level query text instruction $\mathbf{x}_t$ for which reasoning with world knowledge is required, and an input video $\mathbf{x}_v$, 
we aim to build a model $\varphi_{\theta}(\cdot)$ that outputs a binary mask sequence $\mathcal{M}$ representing the described object in the input video:
\begin{align}
   \mathcal{M} = \varphi_{\theta}(\mathbf{x}_t, \mathbf{x}_v),
\end{align}
where the input video $\mathbf{x}_v = \{f_t\}_{t=1}^{T} \in \mathbb{R}^{T \times H \times W \times 3}$ contains $T$ frames, 
and each frame $f_t$ has a size of $H \times W$.
The output binary mask sequence $\mathcal{M} = \{m_t\}_{t=1}^T \in \mathbb{R}^{T \times H \times W}$ has the same frame number and size.

ReasonVOS shares a similar formulation of input and output with the Referring VOS~\cite{seo2020urvos} but is far more challenging.
The principal difference stems from the complexity of the query text in ReasonVOS. 
Unlike simple phrases in Referring VOS that directly describe the appearance, action, or localization characteristics (e.g., \emph{``the running car''}), the query texts in ReasonVOS involve more complex expressions that require world knowledge and common sense (e.g., \emph{``the car powered by electricity''}) or expressions that require complex understanding and inference about video content and upcoming events (e.g., \emph{``Which car is most likely to win the race?''}).

\subsection{Architecture of VISA}
\noindent \textbf{Overview.} 
As shown in Fig.~\ref{fig:architecture}, VISA consists of three main components, namely Text-guided Frame Sampler, multi-modal Large Language Model~(LLM), and Object Tracker.
Specifically, (a) the input video $\mathbf{x}_v$ is fed to the Text-guided Frame Sampler~(TFS), which outputs a target frame $f_{tgt}$ for segmentation and $T_r$ corresponding reference frames $\mathbf{x}_r$ to gain long-term information, guided by the text instructions $\mathbf{x}_t$. 
(b) Then the selected frames are fed into the multi-modal LLM, generating text output including a special token $<$\textsc{Seg}$>$ for segmentation of the target frame $f_{tgt}$. 
(c) Finally, an Object Tracker is utilized to generate the segmentation masks of all frames $\mathcal{M}$ via bi-directional mask propagation.

\noindent \textbf{Text-guided Frame Sampler.} 
%\weidi{can we give high-level goal first, something like, given an input video $\mathbf{x}_v = \{I_1, \dots, I_n\}$, and a text query .... our goal is to train a frame sampler, that outputs .... }
%\weidi{then say, adopt a ViT-based visual encoder, namely, ....it first .... then ....}
Given input video $\mathbf{x}_v$ comprising $T$ frames where each frame is represented by $L$ visual tokens, the total number of tokens to be processed by the multi-modal LLM is $T \times L$.
For segmentation purposes, $L$ should be large enough to maintain spatial details, instead of pooling into a few tokens such as in Video-ChatGPT~\cite{maaz2023video}.
Consequently, it is computationally intractable to directly feed such numerous visual tokens to the multi-modal LLM.
 
As shown in Fig.~\ref{fig:architecture}, the text query \emph{``Which person will take the baton?''} could be answered within the last few frames of the video, while the rest frames are irrelevant to the question.
Inspired by this, we adopt LLaMA-VID~\cite{li2023llama}, a multi-modal LLM that abstracts each input frame into two visual tokens and enables long video processes, to serve as a Text-guided Frame Sampler~(TFS).
TFS generates the most distinguishing frame $f_{tgt}$ and corresponding reference frames $\mathbf{x}_r$ for identifying the described object.
Specifically, a task-specific template is designed: 
\emph{``$<$\textsc{Video}$>$ To find $\{$description$\}$, which percentage mark of the video should I check? Please respond with a number between 0\% and 100\%.''}
We extract the percentage values $p_i$ in the top $K$ responses and use the average value to obtain the target frame $f_{tgt} = f_{T/K \sum p_i}$.
$K$ is set to 10 in this work.
Based on $f_{tgt}$, $T_r$ frames are sampled as reference frames $\mathbf{x}_r$ to obtain long-term temporal correspondence and help with the segmentation of the described object in frame $f_{tgt}$.
We adopt multiple reference sampling strategies in this work, such as Local sampling and Global sampling.
The details and ablation studies of different reference sampling strategies are shown in Ablation Study Sec.~\ref{sec:ablations}.

\noindent \textbf{Multi-Modal Large Language Model.}
Each frame in $\mathbf{x}_r$ and $f_{tgt}$ are encoded via ViT~\cite{dosovitskiy2020image} and tokenized into $L$ visual embeddings by Spatial Merging~\cite{jin2023chat}, yielding visual tokens $<\!\mathbf{x}_r\!>$ and $<\!f_{tgt}\!>$.
Then, the concatenated visual and text tokens are fed to a Multi-Modal LLM to generate the text output containing a special token $<$\textsc{Seg}$>$. The task-specific template is designed as: 
\emph{``USER: $<\!f_{tgt}\!>$ $<\!\mathbf{x}_r\!>$
Can you segment the \{description\}? 
ASSISTANT: Yes, it is $<$\textsc{Seg}$>$.''}, where \{description\} will be replaced by the text description, and the text will be tokenized before being fed to multi-modal LLMs.
We extract the last-layer embedding corresponding to the $<$\textsc{Seg}$>$ token and apply an MLP projection layer to generate $h_{seg}$, which serves as the prompt embedding in SAM decoder~\cite{kirillov2023segment}.

Simultaneously, the vision backbone $\mathcal{E}_{\text{v}}$ extracts the visual features of target frame $f_{tgt}$, which is utilized along with the prompt embedding $h_{seg}$ to produce the segmentation mask $m_{tgt}$:
\begin{align}
   m_{tgt} = \text{SAM}(\mathcal{E}_{\text{v}}(f_{tgt}), h_{seg}).
\end{align}
Finally, an Object Tracking method~\cite{cheng2022xmem} is adopted to propagate $m_{tgt}$ bidirectionally to all rest frames and obtain the mask sequence $\mathcal{M}$:
\begin{align}
   \mathcal{M} = \{m_t\}_{t=1}^T = \text{OT}(m_{tgt}, \mathbf{x}_v).
\end{align}

\noindent \textbf{Training.} 
Following LISA~\cite{lai2023lisa}, our model is trained end-to-end using the standard text generation loss $\mathcal{L}_{\text{txt}}$ and the segmentation mask loss $\mathcal{L}_{\text{mask}}$.
The overall objective $\mathcal{L}$ is the weighted sum of $\mathcal{L}_{\text{txt}}$ and $\mathcal{L}_{\text{mask}}$:
\begin{align}
   \mathcal{L} = \lambda_{\text{txt}} \mathcal{L}_{\text{txt}} + \lambda_{\text{mask}} \mathcal{L}_{\text{mask}}.
\end{align}

Specifically, $\mathcal{L}_{\text{txt}}$ is the auto-regressive cross-entropy loss for text generation, and $\mathcal{L}_{\text{mask}}$ is the combination of per-pixel binary cross-entropy (BCE) loss and DICE loss~\cite{milletari2016v}, with corresponding loss weights $\lambda_{\text{bce}}$ and $\lambda_{\text{dice}}$. 
Given the ground-truth targets ($\hat{\mathbf{y}}_{\text{txt}}$, $\hat{m}_{tgt}$) and the predictions $(\mathbf{y}_{\text{txt}}$, $m_{tgt}$), $\mathcal{L}_{\text{txt}}$ and $\mathcal{L}_{\text{mask}}$ can be formulated as:
\begin{align}
   \mathcal{L}&_{\text{txt}} \hspace{7pt} = \text{CE}(\hat{\mathbf{y}}_{\text{txt}}, \mathbf{y}_{\text{txt}}), \\
   \mathcal{L}&_{\text{mask}} = \lambda_{\text{bce}}\text{BCE}(\hat{m}_{tgt}, m_{tgt}) + \lambda_{\text{dice}} \text{DICE}(\hat{m}_{tgt}, m_{tgt}).
\end{align}

\begin{figure*}[!t]
\begin{center}
\includegraphics[width=0.97\linewidth]{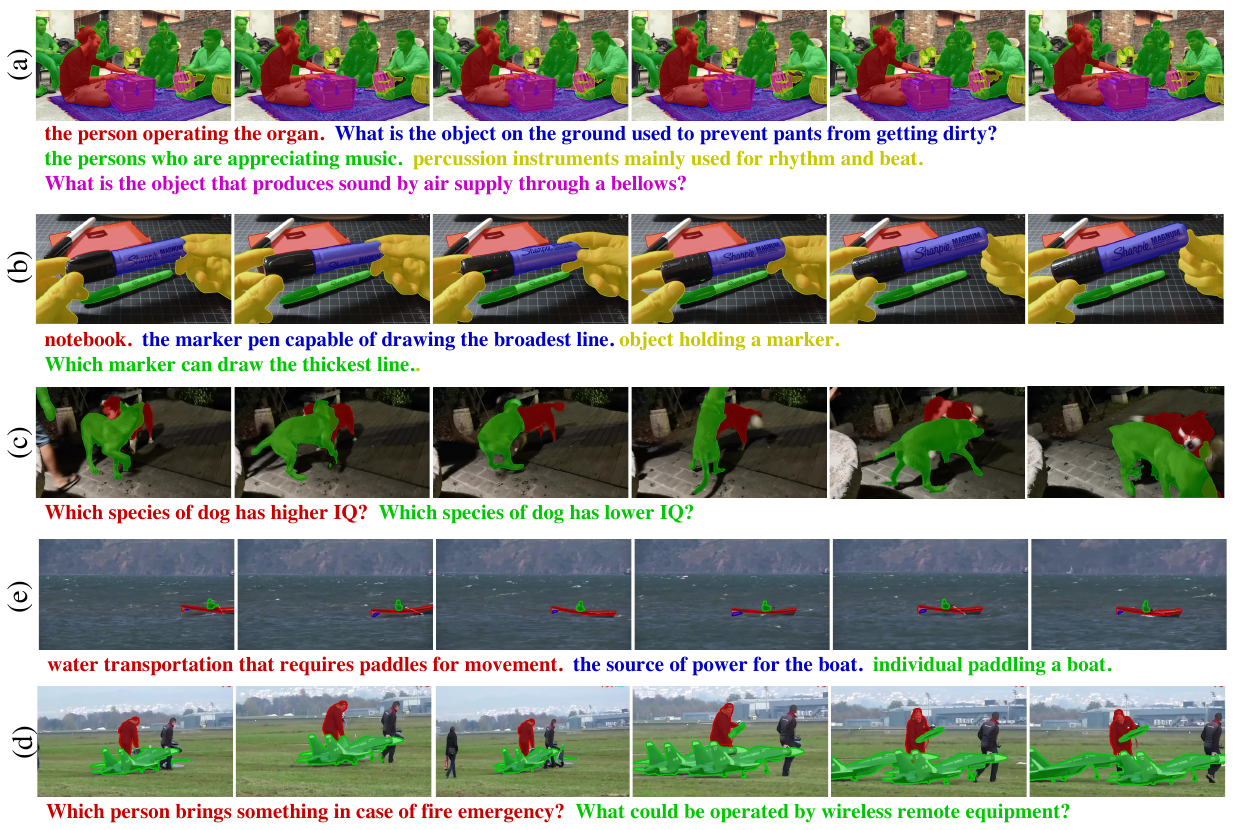}
\end{center}
\vspace{-5mm}
\caption{Visualizations of VISA on ReVOS dataset.}
\label{fig:vis}
\vspace{-3mm}
\end{figure*}

\subsection{ReVOS Dataset}

For the quantitative evaluation of ReasonVOS, it is essential to establish a benchmark characterized by implicit object descriptions and high-quality mask sequences.
To this end, we collect ReVOS, a dataset containing complex text instructions and corresponding high-quality masks in videos for both instruction tuning and evaluation of ReasonVOS.
To guarantee reliable assessment, we collect a diverse set of videos from LV-VIS~\cite{wang2023towards,wang2024ov}, MOSE~\cite{ding2023mose}, OVIS~\cite{qi2021occluded}, TAO~\cite{dave2020tao} and UVO~\cite{wang2021unidentified}.
Subsequently, we annotate the objects in the videos with complex text instructions and match these instructions with the corresponding target mask sequences.

\noindent \textbf{Dataset Statistics}
Overall, our dataset comprises a total of \red{35,074} object-instruction pairs from \red{1,042} videos.
All the videos are divided into a training~(instruction tuning) set and a validation set containing \red{626} videos and \red{416} videos respectively.
The text instructions consist of (1) \red{14,678} implicit descriptions requiring world knowledge and video content reasoning and inference to evaluate the ReasonVOS; (2) \red{20,071} explicit descriptions to evaluate the generalization ability in traditional Referring VOS task; (3) \red{325} descriptions of nonexistent objects for the hallucination evaluation.
Check \url{https://github.com/cilinyan/ReVOS-api} for more detailed dataset information.

\noindent \textbf{Evaluation Metrics}
We follow most previous works on Referring VOS~\cite{seo2020urvos, ding2023mevis} to adopt $\mathcal{J}\&\mathcal{F}$ as the main evaluation metric, which is the average of region similarity $\mathcal{J}$ and contour accuracy $\mathcal{F}$.
As for the evaluation of hallucination, we adopt the robustness score $\mathcal{R}$ introduced in R2VOS~\cite{li2022r}.

%% file: section/4-experiment.tex
\section{Experiments}

\subsection{Dataset}
\noindent \textbf{Training Dataset.}
Our training data consists of vanilla Referring VOS datasets, Video Question-Answering datasets, Image datasets, and the ReVOS dataset. 
The details are as follows: (1) Referring VOS datasets.
We use Ref-YouTube-VOS~\cite{seo2020urvos}, MeViS~\cite{ding2023mevis}, and Ref-DAVIS17~\cite{pont20172017} during training to learn the projections between objects in videos and text expressions.
Those Referring VOS datasets provide input videos, explicit short descriptions, and corresponding object masks. 
(2) Video Question-Answering datasets. 
To achieve better reasoning and question-answering ability in videos of the multi-modal LLM, we include the video instruction data from Video-ChatGPT~\cite{maaz2023video}. 
The answer template ``It's $<$\textsc{Seg}$>$'' is replaced by the original annotated answers in those datasets, and the corresponding segmentation loss is ignored during training.
(3) Image datasets. Images could be regarded as one-frame videos. 
Thereby, we adopt all the vanilla datasets used by LISA~\cite{lai2023lisa} in our work to achieve more stable training.
(4) ReVOS dataset. The above-mentioned training datasets contain no ReasonVOS samples.
Therefore, we include the ReVOS dataset during training to achieve more comprehensive reasoning and object segmentation ability in videos. 
The implementation details are shown in Supplementary Material.

\noindent \textbf{Evaluation Dataset}
We evaluate VISA on both Video datasets and Image datasets.
(1) Video datasets. We use the ReVOS dataset to evaluate the performance of ReasonVOS; we use Ref-YouTube-VOS~\cite{seo2020urvos}, MeViS~\cite{ding2023mevis}, and Ref-DAVIS17~\cite{pont20172017} to evaluate the performance of vanilla Referring VOS.
(2) Image datasets. We use ReasonSeg~\cite{lai2023lisa}, refCOCO~\cite{kazemzadeh2014referitgame}, refCOCO+~\cite{kazemzadeh2014referitgame}, and refCOCOg~\cite{mao2016generation} to evaluate the generalization ability of VISA on image-level segmentation tasks.

\subsection{Implementation Details}
We adopt the decoder in SAM~\cite{kirillov2023segment} as the segmentation decoder. 
We choose Chat-UniVi-7B and Chat-UniVi-13B~\cite{jin2023chat} as the pre-trained Multi-Modal LLM.
The number of visual tokens $L$ per frame is set to 112, the same as the number in Chat-UniVi.
We utilize XMem~\cite{cheng2022xmem}, a semi-supervised Video Object Segmentation method as the Object Tracker.
The Text-guided Frame Sampler, Visual Backbone, and Object Tracker are all frozen during the training.
Only the multi-modal LLM and SAM decoders are trainable.
We leverage LoRA~\cite{hu2021lora} to perform efficient fine-tuning of the multi-modal LLM.
During training, we randomly sample a target frame $f_{tgt}$ and 8-12 reference frames $\mathbf{x}_t$ per video instead of using the Text-guided Frame Sampler, to achieve more comprehensive training.
During inference, we use the Text-guided Frame Sampler to obtain $f_{tgt}$ and 12 reference frames $\mathbf{x}_t$ with the Global-Local sampling strategy.
We use 8 NVIDIA 80G A100 GPUs for training. 
The training scripts are based on the deepspeed~\cite{rasley2020deepspeed} engine.
We train VISA for 10 epochs with a batch size of 128.
Specifically, the batch size per device is set to 1, and the gradient accumulation step is set to 16, leading to 128 samples on 8 GPUs in total.
We employ the AdamW~\cite{loshchilov2017decoupled} optimizer with a cosine schedule.
The learning rate is set to 2e-5.
All input frames are resized to 224 $\times$ 224 before feeding into the LLM, while the frame for the segmentation branch is resized to 1024 $\times$ 1024.
The weights of the text generation loss $\lambda_{\text{txt}}$ and the mask loss $\lambda_{\text{mask}}$ are set to 1.0 and 1.0, respectively.
The weights of the binary cross-entropy loss $\lambda_{\text{bce}}$ and the dice loss $\lambda_{\text{dice}}$ are set to 2.0 and 0.5, respectively.
Check \url{https://github.com/cilinyan/VISA} for more implementation details.

\subsection{Comparison}

\begin{table*}[!t]
\caption{Performance comparison on ReVOS dataset. $*$ means the method is reproduced in this work. (IT) means instruction tuning with the ReVOS training set. $\mathcal{R}$ is the robustness score.}
\vspace{-2mm}
\centering
\setlength{\tabcolsep}{5pt}
\begin{tabular}{l|l|cccccccccc}
\toprule
\multirow{2}{*}{Method}&\multirow{2}{*}{Backbone} & \multicolumn{3}{c}{referring} & \multicolumn{3}{c}{reasoning} & \multicolumn{3}{c}{overall} & \multirow{2}{*}{$\mathcal{R}$} \\ \cmidrule(lr){3-5} \cmidrule(lr){6-8} \cmidrule(lr){9-11}
                      & & $\mathcal{J}$       & $\mathcal{F}$       & $\mathcal{J\&F}$      & $\mathcal{J}$       & $\mathcal{F}$       & $\mathcal{J\&F}$      & $\mathcal{J}$       & $\mathcal{F}$       & $\mathcal{J\&F}$      &       \\ \midrule
ReferFormer~\cite{wu2022language}&Resnet50&16.6&17.1&16.9&11.9&13.8&12.8&14.3&15.4&14.9&4.9\\
MTTR~\cite{botach2022end}&Video-Swin-T&29.8&30.2&30.0&20.4&21.5&21.0&25.1&25.9&25.5&5.6\\
LMPM~\cite{ding2023mevis}&Swin-T&29.0&39.1&34.1&13.3&24.3&18.8&21.2&31.7&26.4&3.2\\
ReferFormer~\cite{wu2022language}&Video-Swin-B&31.2&34.3&32.7&21.3&25.6&23.4&26.2&29.9&28.1&8.8\\
\makecell[l]{LLaMA-VID~\cite{li2023llama}+LMPM}&Swin-T&29.0&39.1&34.1&12.8&23.7&18.2&20.9&31.4&26.1&3.4\\
\midrule
\midrule
LISA~\cite{lai2023lisa}&LLaVA-7B&44.3 & 47.1 & 45.7 & 33.8 & 38.4 & 36.1 & 39.1 & 42.7 & 40.9 & 9.3 \\
LISA*~\cite{lai2023lisa}&LLaVA-13B&45.2 & 47.9 & 46.6 & 34.3 & 39.1 & 36.7 & 39.8 & 43.5 & 41.6 & 8.6 \\
TrackGPT(IT)*~\cite{stroh2024trackgpt}&LLaVA-7B&46.7 & 49.7 & 48.2 & 36.8 & 41.2 & 39.0 & 41.8 & 45.5 & 43.6 & 11.6 \\
TrackGPT(IT)*~\cite{stroh2024trackgpt}&LLaVA-13B&48.3 & 50.6 & 49.5 & 38.1 & 42.9 & 40.5 & 43.2 & 46.8 & 45.0 & 12.8 \\ \midrule
VISA&Chat-UniVi-7B&51.1&54.7&52.9&36.7&41.7&39.2&43.9&48.2&46.1&7.9\\
VISA&Chat-UniVi-13B&52.3&55.8&54.1&38.3&43.5&40.9&45.3&49.7&47.5&8.3\\
VISA(IT)&LLaVA-7B&49.4&52.6&51.0&40.5&45.8&43.2&44.9&49.2&47.1&\underline{15.3}\\
VISA(IT)&LLaVA-13B&\textbf{55.7}&\underline{59.0}&\textbf{57.4}&\underline{41.9}&\underline{46.5}&\underline{44.2}&\textbf{48.8}&\underline{52.8}&\underline{50.8}&15.1\\
VISA(IT)&Chat-UniVi-7B&49.2&52.6&50.9&40.6&45.4&43.0&44.9&49.0&46.9&\textbf{15.5}\\
VISA(IT)&Chat-UniVi-13B&\underline{55.6}&\textbf{59.1}&\textbf{57.4}&\textbf{42.0}&\textbf{46.7}&\textbf{44.3}&\textbf{48.8}&\textbf{52.9}&\textbf{50.9}&14.5\\
\bottomrule

\end{tabular}
\label{tab:reasoning}
\end{table*}

\begin{table*}[!t]
\caption{Performance comparison on Referring VOS datasets. The results on MeViS above the horizontal line are provided in LMPM~\cite{ding2023mevis}, which are all obtained with the Swin-T backbone. The results of TrackGPT on MeViS are generated by our reproduced model.}
\vspace{-2mm}
\centering
\setlength{\tabcolsep}{7.8pt}
\begin{tabular}{l|l|ccccccccc}
\toprule
\multirow{2}{*}{Methods} &\multirow{2}{*}{Backbone}& \multicolumn{3}{c}{MeViS} & \multicolumn{3}{c}{Ref-YT-VOS} & 
\multicolumn{3}{c}{Ref-DAVIS17} \\ \cmidrule(lr){3-5} \cmidrule(lr){6-8} \cmidrule(lr){9-11}
                        & & $\mathcal{J}$       & $\mathcal{F}$       & $\mathcal{J\&F}$     & $\mathcal{J}$       & $\mathcal{F}$       & $\mathcal{J\&F}$     &$\mathcal{J}$       & $\mathcal{F}$       & $\mathcal{J\&F}$  \\ \midrule
URVOS~\cite{seo2020urvos}&ResNet50&25.7 &29.9 &27.8&45.3 &49.2 &47.2 &47.3 &56.0 &51.6 \\
LBDT~\cite{ding2022language}&ResNet50&27.8 &30.8 &29.3 &48.2 &50.6 &49.4 &-&-&54.1 \\
MTTR~\cite{botach2022end}&Video-Swin-T&28.8 &31.2 &30.0 &54.0 &56.6 &55.3 &-&-&-\\
ReferFormer~\cite{wu2022language}&Video-Swin-B&29.8 &32.2 &31.0 &61.3&64.6 &62.9 &58.1 &64.1 &61.1 \\
LMPM~\cite{ding2023mevis}&Swin-T&34.2 &40.2 &37.2 &-&-&-&-&-&-\\
OnlineRefer~\cite{wu2023onlinerefer}&Swin-L&-&-&-&\textbf{61.6} &\textbf{65.5} &\textbf{63.5} &61.6 &67.7 &64.8 \\\midrule 
LISA~\cite{lai2023lisa}&LLaVA-7B&35.1&39.4&37.2&53.4 &54.3 &53.9 &62.2&67.3&64.8\\
LISA~\cite{lai2023lisa}&LLaVA-13B&35.8&40.0&37.9&54.0 &54.8 &54.4 &63.2&68.8&66.0\\
TrackGPT~\cite{stroh2024trackgpt}&LLaVA-7B&37.6&42.6&40.1&55.3&57.4&56.4&59.4&67.0&63.2 \\
TrackGPT~\cite{stroh2024trackgpt}&LLaVA-13B&39.2&43.1&41.2&58.1&60.8&59.5&62.7&70.4&66.5\\
VISA~(Ours)&Chat-UniVi-7B&\underline{40.7}&\underline{46.3}&\underline{43.5}&59.8&63.2&61.5&\underline{66.3}&\underline{72.5}&\underline{69.4}\\
VISA~(Ours)&Chat-UniVi-13B&\textbf{41.8}&\textbf{47.1}&\textbf{44.5}&\underline{61.4}&\underline{64.7}&\underline{63.0}&\textbf{67.0}&\textbf{73.8}&\textbf{70.4} \\
\bottomrule
\end{tabular}
\label{tab:referring}
\end{table*}

\noindent \textbf{ReVOS.}
The results comparison on ReVOS are shown in Tab.~\ref{tab:reasoning}.
Compared with traditional methods~(even with extremely large visual backbones), our proposed VISA(IT)-7B generally achieves over \red{20} $\mathcal{J}\&\mathcal{F}$ improvements in terms of reasoning.
Those traditional works are limited to short explicit references and have no capability of reasoning and understanding the implicit text queries.

VISA(IT)-7B outperforms the single frame method LISA-7B~\cite{lai2023lisa} by \red{6.0} $\mathcal{J}\&\mathcal{F}$ in terms of overall performance, which indicates the ability of VISA to conduct video-level segmentation.
Recent work TrackGPT~\cite{stroh2024trackgpt} incorporates tracking with LLMs, yet the multi-modal LLMs in TrackGPT only process a single frame at one time, leading to poor temporal information gathering.
As a consequence, VISA(IT)-7B outperforms TrackGPT(IT)-7B by \red{3.3} $\mathcal{J}\&\mathcal{F}$ overall.
Moreover, as we include plenty of negative samples~(text queries of nonexistent objects) in the ReVOS training set, the hallucination of VITA is much lower than in existing methods. As shown, the robustness scores $\mathcal{R}$ of VISA are much higher than existing methods.

VQA+RerferringVOS could serve as a baseline model, 
but can not solve ReasonVOS well.
As suggested in Tab.~\ref{tab:reasoning}, we use LLaMA-VID, a Video-VQA method, to transfer the complex questions~(e.g., scared dog) into low-level descriptions~(e.g., dog on left), and then employ LMPM, a RVOS method, to segment the described objects.
As shown below, LLaMA-VID + LMPM performs worse than LMPM on ReVOS reasoning set. That is because the existing video-VQA methods take only a few visual tokens per frame, which is too vague to localize the described objects and brings mistakes when converting complex questions into low-level expressions. 

Note that VISA with LLaVA-7B~\cite{liu2024visual} and Chat-UniVi-7B~\cite{jin2023chat} achieve similar performance.
Chat-UniVi could process a flexible number of visual tokens via spatial merging, thus we chose it in this work.
We visualize the results of VISA on the ReVOS dataset in Fig.~\ref{fig:vis}.

\noindent \textbf{Referring VOS.}
To demonstrate that VISA generalizes well in the vanilla Referring VOS task, we compare VISA with the existing methods in Tab.~\ref{tab:referring}.
As shown, VISA achieves the SOTA results over three widely used Referring VOS datasets.
%Specifically, VISA-13B outperforms ReferFormer~\cite{wu2022language} with strong Video-Swin-B backbone by 13.6\%, 0.1\%, and 9.2\% $\mathcal{J}\&\mathcal{F}$ on MeVIS, Ref-YT-VOS, and Ref-DAVIS17 respectively.
%Ref-YT-VOS contains relatively fewer objects per video and the target objects are usually .

\begin{table*}[t]
\caption{Performance comparison on Image Segmentation datasets.
% $^{\dagger}$ denotes the results obtained from the model we trained using the official GitHub repository
% ~\url{https://github.com/dvlab-research/LISA}
% .
$^{\dagger}$ denotes the results obtained from the model we trained via LISA's official GitHub repository.
}
\vspace{-2mm}
\centering
\scalebox{1}{
\begin{tabular}{l|l|cccccccc|cc}
\toprule
\multirow{2}{*}{Methods} &\multirow{2}{*}{Backbone}&\multicolumn{3}{c}{refCOCO} & \multicolumn{3}{c}{refCOCO+} & 
\multicolumn{2}{c|}{refCOCOg}&\multicolumn{2}{c}{ReasonSeg}\\ \cmidrule(lr){3-5} \cmidrule(lr){6-8} \cmidrule(lr){9-10} \cmidrule(lr){11-12} 
&&val&testA&testB&val&testA&testB&val(U)&test(U)&gIoU&cIoU\\ \midrule
MCN~\cite{luo2020multi}&Darknet53&62.4&64.2&59.7&50.6&55.0&44.7&49.2&49.4&-&-\\
VLT~\cite{ding2021vision}&Darknet53&65.7&68.3&62.7&55.5&59.2&49.4&53.0&56.7&-&-\\
CRIS~\cite{wang2022cris}&ResNet101&70.5&73.2&66.1&62.3&68.1&53.7&59.9&60.4&-&-\\
LAVT~\cite{yang2022lavt}&Swin-B&72.7&75.8&68.8&62.1&68.4&55.1&61.2&62.1&-&-\\
ReLA~\cite{liu2023gres}&Swin-B&73.8&76.5&70.2&66.0&71.0&57.7&65.0&66.0&-&-\\
X-Decoder~\cite{zou2023generalized}&DaViT-L&-&-&-&-&-&-&64.6&-&22.6&17.9\\
SEEM~\cite{zou2024segment}&DaViT-L&-&-&-&-&-&-&65.7&-&25.5&21.2\\ \midrule
LISA~\cite{lai2023lisa}&LLaVA-7B&
74.9&79.1&72.3&65.1&70.8&58.1&67.9&70.6&52.9&54.0\\
LISA$^{\dagger}$&LLaVA-7B&
70.8&73.7&66.3&58.1&63.2&51.2&63.8&64.8&48.1&53.7\\
VISA~(Ours)&Chat-UniVi-7B&72.4&75.5&68.1&59.8&64.8&53.1&65.5&66.4&52.7&57.8\\
\bottomrule
\end{tabular}
}
\label{tab:image}
\end{table*}

\noindent \textbf{Image Datasets.} Images could be regarded as single-frame videos. 
Therefore, VISA could be directly applied to image datasets without any modification.
As shown in Tab.~\ref{tab:image}, VISA achieves comparable performance with LISA on three referring image segmentation datasets, while significantly outperforming traditional methods by over 20\% on the ReasonSeg~\cite{lai2023lisa} dataset.
The results indicate that VISA has a strong generalization ability on vanilla referring image segmentation and reasoning image segmentation.

\begin{table*}[t]
\caption{Performance on ReVOS validation set with different training datasets. The columns with $\checkmark$ mean the corresponding datasets are adopted during training.}
\vspace{-2mm}
\centering
\begin{tabular}{c|c|c|c|cccccc}
\toprule
\multirow{2}{*}{ReferringVOS} & \multirow{2}{*}{VQA} & \multirow{2}{*}{Image} & \multirow{2}{*}{ReVOS}&\multicolumn{3}{c}{referring} & \multicolumn{3}{c}{reasoning} \\ \cmidrule(lr){5-7} \cmidrule(lr){8-10}
&&&& $\mathcal{J}$ & $\mathcal{F}$& $\mathcal{J\&F}$ & $\mathcal{J}$  & $\mathcal{F}$ & $\mathcal{J\&F}$\\ \midrule
&\checkmark&\checkmark&\checkmark&45.9&49.3&47.6&37.4&42.3&39.9\\
\checkmark&&\checkmark&\checkmark&48.6&52.1&50.3&38.9&43.9&41.4\\
\checkmark&\checkmark&&\checkmark&32.3&36.2&34.2&30.9&36.1&33.5\\
\checkmark&\checkmark&\checkmark&&51.1&54.7&52.9&36.7&41.7&39.2\\
\checkmark&\checkmark&\checkmark&\checkmark&49.2&52.6&50.9&40.6&45.4&43.0\\
\bottomrule
\end{tabular}
\label{tab:trainingdata}
\end{table*}

\subsection{Ablation Studies}
\label{sec:ablations}
\noindent \textbf{Training Datasets}
In Tab.~\ref{tab:trainingdata}, we show the contribution of each type of dataset during training. 
As shown, without Referring VOS datasets, the performance drops by \red{3.3}\%  $\mathcal{J}\&\mathcal{F}$ and \red{3.1}\% $\mathcal{J}\&\mathcal{F}$ in terms of referring and reasoning segmentation, respectively.
That is because Referring VOS datasets provide text expressions and mask sequence pairs in videos, aligning the video domain and linguistic domain, thus generally helping with the vanilla referring segmentation and reasoning segmentation tasks in videos.
Without Image datasets, the performance of VISA significantly drops by \red{16.7}\% and \red{9.5}\%.
Generally, image datasets have much larger scales than video datasets, leading to more robust feature alignment and stronger generalization ability.
The models tend to be overfitting during training without image datasets.
By instruction tuning on ReVOS, VISA further gains \red{3.8}\% $\mathcal{J}\&\mathcal{F}$ performance improvements of reasoning segmentation, while the performance of referring segmentation barely changes, which shows the effectiveness of our collected ReVOS dataset to improve the complex video reasoning ability.

\begin{table}[]
    \centering
    \setlength{\tabcolsep}{4pt}
    \caption{Overall $\mathcal{J}\&\mathcal{F}$ on ReVOS with different number $T_r$ of reference frames $\mathbf{x}_r$ and different sampling strategies.}
    \vspace{-2mm}
    \begin{tabular}{l|c|c|ccc}
        \toprule
        &$T_r$ &w/o Sample& Global & Local & Global-Local \\ 
         \midrule
        \multirow{3}{*}{$f_0$}&0&42.6&-&-&-\\
        &6&-&43.9&44.5&44.6\\
        &12&-&44.5&44.9&45.0\\ \midrule
        \multirow{3}{*}{$f_{tgt}$}&0&44.3&-&-&-\\
        &6&-&46.0&46.1&46.3\\
        &12&-&46.7&46.3&46.9\\
        \bottomrule
    \end{tabular}
    \label{tab:frameasample}
\end{table}

\begin{table}[]
    \centering
    \caption{The performance comparison on ReVOS with different number $L$ of visual tokens per frame.}
    \vspace{-2mm}
    \setlength{\tabcolsep}{2.5pt}
    \begin{tabular}{l|l|cccccc}
        \toprule
        \multirow{2}{*}{L}&\multirow{2}{*}{backbone}&\multicolumn{3}{c}{referring} & \multicolumn{3}{c}{reasoning}\\ \cmidrule(lr){3-5} \cmidrule(lr){6-8}
        && $\mathcal{J}$ & $\mathcal{F}$& $\mathcal{J\&F}$ & $\mathcal{J}$  & $\mathcal{F}$ & $\mathcal{J\&F}$\\
        \midrule
        256&LLaVA-7B&49.4&52.6&51.0&40.5&45.8&43.2\\
        112&Chat-UniVi-7B&49.2&52.6&50.9&40.6&45.4&43.0\\
        56&Chat-UniVi-7B&44.9&48.5&46.7&36.5&41.5&39.0\\
        \bottomrule
    \end{tabular}
    \label{tab:llm}
\end{table}

\begin{figure*}[!t]
\begin{center}
\includegraphics[width=1.0\linewidth]{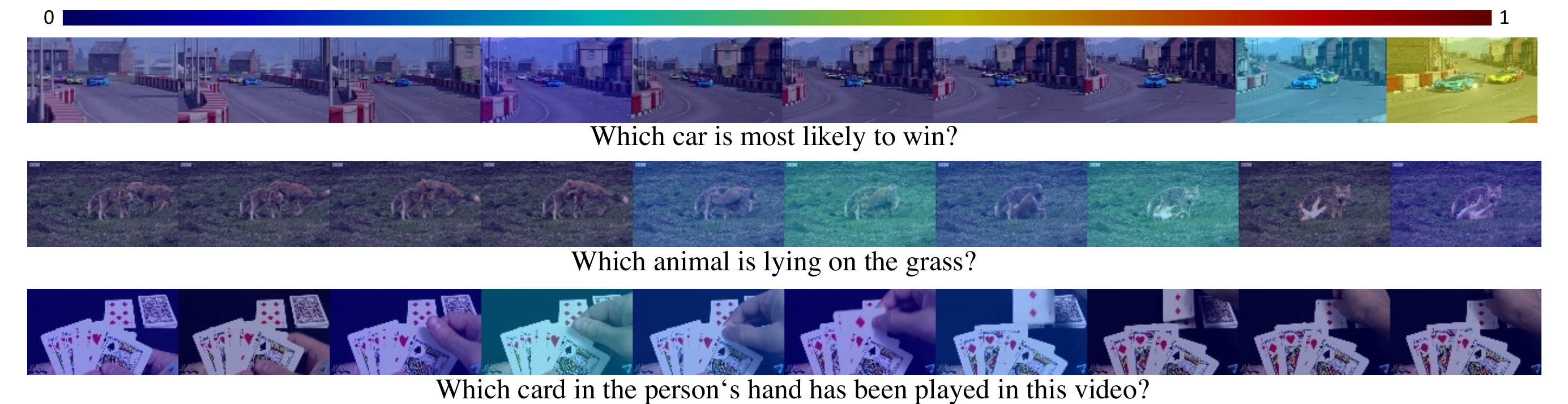}
\end{center}
\vspace{-5mm}
\caption{Heatmaps of the target frame $f_{tgt}$. To draw the heatmap, we generate 10 responses with the Text-guided Frame Sampler~(TFS) and obtain the normalized distribution. As shown, the highlighted frames are related to the text queries.}
\label{fig:heatmap}
\vspace{-2mm}
\end{figure*}

\noindent \textbf{Target Frame $f_{tgt}$.}
We visualize the heatmaps of the target frame $f_{tgt}$ in Fig.~\ref{fig:heatmap}.
As shown in the figure, the frames related to given text expressions are highlighted.
We show the ablation of $f_{tgt}$ in Tab.~\ref{tab:frameasample}. $f_0$ means we directly segment the object in the first frame and propagate to the rest frames, while $f_{tgt}$ means we use TFS to obtain the target frame $f_{tgt}$ and segment $f_{tgt}$ in consequence.
As shown, the performance with $f_{tgt}$ generally outperforms $f_0$ by around \red{2}\% under different settings, which indicates the effectiveness of TFS in obtaining the significant moments related to described objects.

\noindent \textbf{Reference Sampling Strategies.} In Tab.~\ref{tab:frameasample}, Global means we uniformly sample frames through the whole video as reference frames $\mathbf{x}_r$, while Local means we sample contiguous frames centered by $f_{tgt}$ as reference frames.
Global-Local means the combination of 
$T_r/2$ frames from Global and $T_r/2$ frames from Local.
As shown, Global-Local sampling slightly outperforms the separate ones, thus we adopt it in VISA.
As the number $T_r$ of reference frames $\mathbf{x}_r$ increases, the performance gradually improves.
To keep feasible training and inference, we adopt $T_r$=12 in VISA.
Overall, with $f_{tgt}$ and Global-Local sampling, VISA achieves \red{4.3}\% $\mathcal{J}\&\mathcal{F}$ improvements on the ReVOS dataset.

\noindent \textbf{Number $L$ of visual tokens.}
The performance comparison under different numbers $L$ of visual tokens per frame is shown in Tab~\ref{tab:llm}.
For $L$=256, we adopt LLaVA-7B~\cite{liu2024visual} as the backbone, which takes 256 visual tokens for the input image.
For $L$=112 and $L$=52, we use Chat-UniVi-7B~\cite{jin2023chat} as the backbone, and utilize the Spatial Merging~\cite{jin2023chat} to project visual tokens to corresponding numbers.
As shown, VISA with 256 tokens and 112 tokens per frame achieves comparable performance on ReVOS.
When $L$ is set to 52, the performance of VISA significantly drops.
Therefore, we adopt $L$=112 in this work.

\subsection{Limitations}
\noindent \textbf{Small Objects} Limited by the number of visual tokens per frame~(for instance, 256 in LISA~\cite{liu2024visual}, and 112 in VISA), the current methods have a poor ability to capture very small objects.
As shown in Fig.~\ref{fig:vis} (d), the small  
paddles are not segmented.
A multi-modal LLM with more input visual tokens could relieve this issue, but will lead to more computational burden and complex training process.

\noindent \textbf{Temporal Information Gathering} In this work, we intuitively adopt a Text-guided Frame Sampler to select a feasible number of important frames for the multi-modal LLM. 
The performance highly relies on the accuracy of located frames.
Some objects may only appear in a few frames, which is hard to locate.
As shown in Fig.~\ref{fig:vis} (e), the person with a fire tank only appears in one frame, while VISA falls to locate this frame and segment another person in consequence.
Moreover, the text description could require extremely long temporal correspondence, but VISA could only handle a few selected frames at the same time. 
To this end, a more effective way to gather long-term temporal information while maintaining spatial details is required.
We leave those issues to our future work.

%% file: section/5-conclusion.tex
\section{Conclusion}
In this work, we propose a new task, ReasonVOS, which aims to generate object mask sequences in response to text queries that require complex reasoning and inference abilities within video contexts.
To tackle ReasonVOS, we design VISA (Video-based large language Instructed Segmentation Assistant), to leverage the world knowledge and reasoning capabilities of multi-modal LLMs while possessing the ability to segment and track objects in videos.
Moreover, we collect a large-scale dataset ReVOS, containing \red{35,074} expression-mask pairs from \red{1,042} videos for the instruction tuning and evaluation of ReasonVOS methods.
Experiments on eight various datasets show that our proposed VISA not only enables the reasoning segmentation ability in videos but also generally provides SOTA performance on traditional video and image segmentation tasks.

%% file: section/6-appendix.tex
\clearpage

\onecolumn
\noindent
{\LARGE \textbf{Appendix}}

\renewcommand\thesection{\Alph{section}}
\setcounter{section}{0}

\section{Visualizations of Annotated Frames}
\begin{figure*}[ht]
\vspace{-1mm}
\begin{center}
%\fbox{\rule{0pt}{1.8in} \rule{0.9\linewidth}{0pt}}
\includegraphics[width=1.0\linewidth]{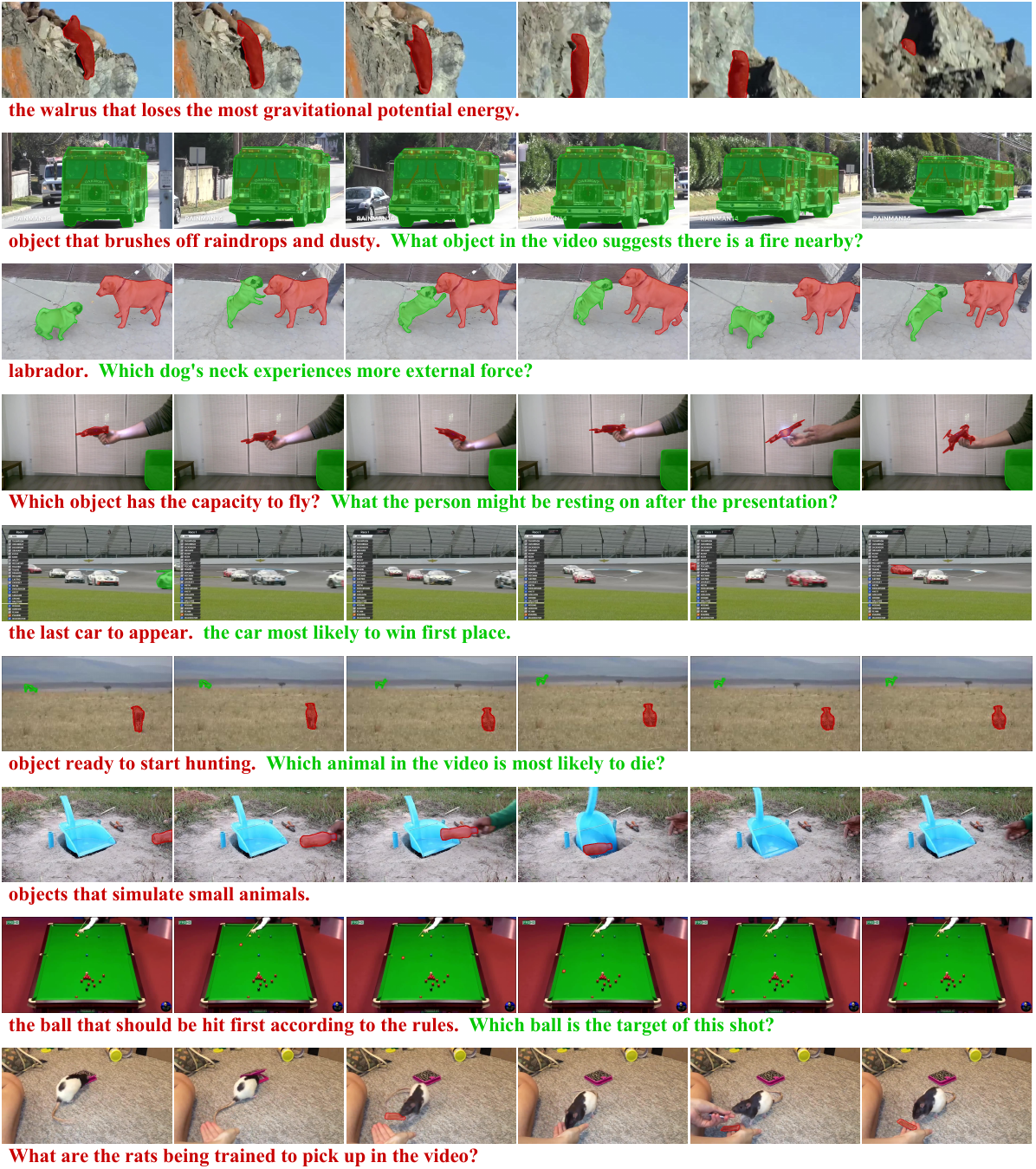}
\end{center}
   \vspace{-5mm}
\caption{Sample videos in ReVOS.}
\vspace{-1mm}
\label{fig:anno_vis}
\end{figure*}

Examples of annotated videos in ReVOS are shown in~\ref{fig:anno_vis}.

